\documentclass[10pt,twocolumn,letterpaper]{article}
\usepackage[table,dvipsnames]{xcolor}
\usepackage{iccv}              
\usepackage{multirow}
\usepackage{makecell}
\usepackage{pifont}
\usepackage{array}
\usepackage{booktabs} 

\usepackage{tabularx}
\usepackage{amsmath}
\usepackage[accsupp]{axessibility}

\definecolor{iccvblue}{rgb}{0.21,0.49,0.74}
\usepackage[pagebackref,breaklinks,colorlinks,allcolors=iccvblue]{hyperref}

\def\paperID{13154} 
\def\confName{ICCV}
\def\confYear{2025}

\title{Measuring the Impact of Rotation Equivariance on Aerial Object Detection}

\author{
	Xiuyu Wu\textsuperscript{1} \quad
	Xinhao Wang\textsuperscript{1} \quad
	Xiubin Zhu\textsuperscript{1}\quad
	Lan Yang\textsuperscript{1} \quad
	Jiyuan Liu\textsuperscript{2} \quad
	Xingchen Hu\textsuperscript{2}\thanks{Corresponding author. \\Code page: \url{https://github.com/Nu1sance/MessDet}} \\
	\textsuperscript{1}Xidian University \quad
	\textsuperscript{2}National University of Defense Technology\\
	{\tt\small \{xiuyuwu, xhaowang\}@stu.xidian.edu.cn, xbzhu@mail.xidian.edu.cn, yanglan@xidian.edu.cn,} \\
	{\tt\small liujiyuan13@nudt.edu.cn, xhu4@ualberta.ca}
}

\begin{document}
\maketitle
\begin{abstract}

Due to the arbitrary orientation of objects in aerial images, rotation equivariance is a critical property for aerial object detectors. However, recent studies on rotation-equivariant aerial object detection remain scarce. Most detectors rely on data augmentation to enable models to learn approximately rotation-equivariant features. A few detectors have constructed rotation-equivariant networks, but due to the breaking of strict rotation equivariance by typical downsampling processes, these networks only achieve approximately rotation-equivariant backbones. Whether strict rotation equivariance is necessary for aerial image object detection remains an open question. In this paper, we implement a strictly rotation-equivariant backbone and neck network with a more advanced network structure and compare it with approximately rotation-equivariant networks to quantitatively measure the impact of rotation equivariance on the performance of aerial image detectors. Additionally, leveraging the inherently grouped nature of rotation-equivariant features, we propose a multi-branch head network that reduces the parameter count while improving detection accuracy. Based on the aforementioned improvements, this study proposes the \textbf{M}ulti-branch head rotation-\textbf{e}quivariant \textbf{s}ingle-\textbf{s}tage \textbf{Det}ector (MessDet), which achieves state-of-the-art performance on the challenging aerial image datasets DOTA-v1.0, DOTA-v1.5 and DIOR-R with an exceptionally low parameter count. 
\end{abstract}    
\section{Introduction}
\label{sec:intro}

\begin{figure}[t]
  \centering
   \includegraphics[width=0.99\linewidth]{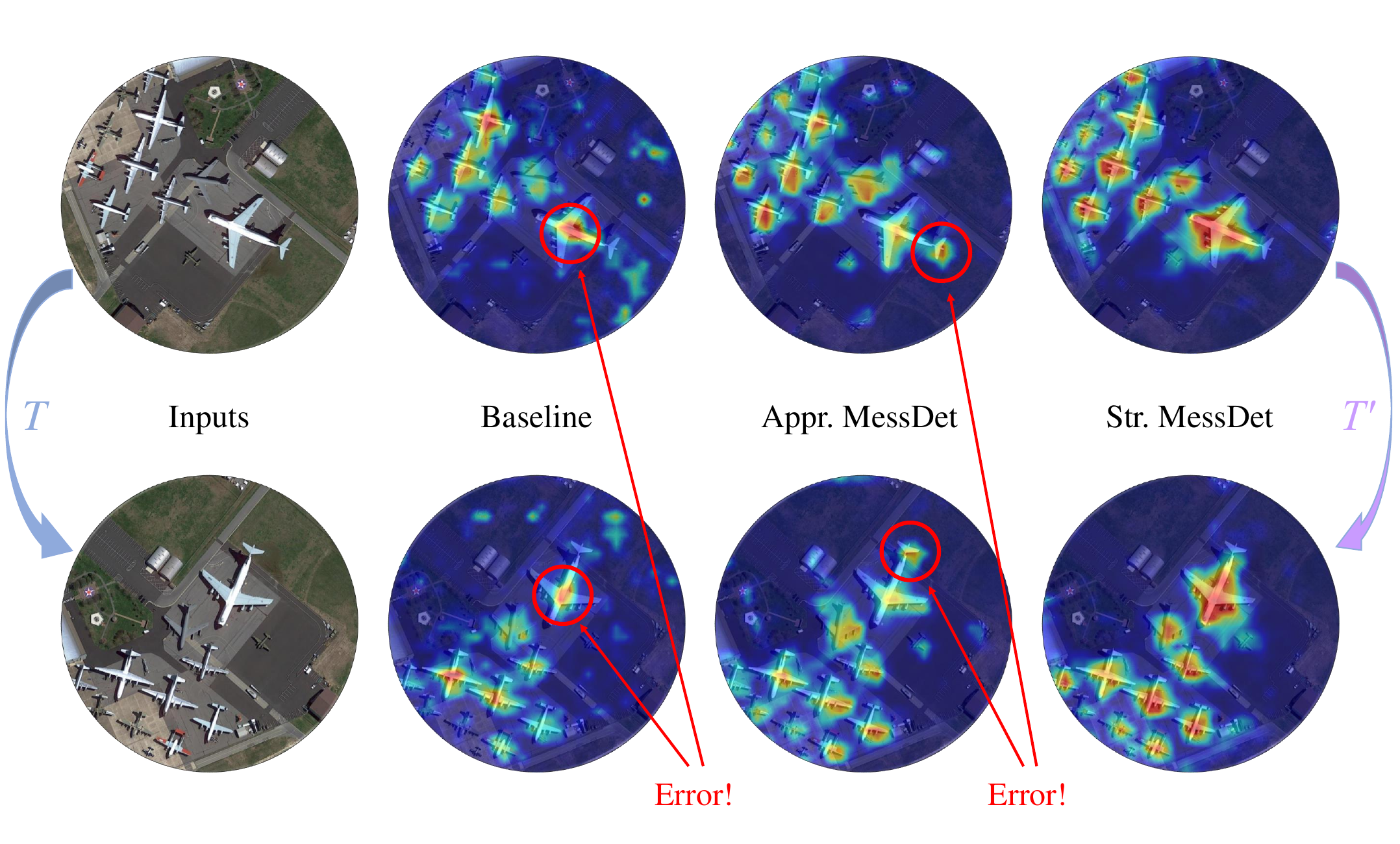}

   \caption{Visualization of features extracted by different methods. When the input is rotated, features extracted by conventional CNNs are easily affected, leading to degraded detection performance. In contrast, our MessDet framework includes both approximately rotation-equivariant (Appr. MessDet) and strictly rotation-equivariant (Str. MessDet) variants. While Appr. MessDet reduces rotation error during training, Str. MessDet eliminates such errors entirely, resulting in more robust detection performance.}
   \label{fig:1}
\end{figure}

The primary difference between aerial object detection \cite{8113128,10168277,cheng2016survey} and general object detection \cite{10028728,Girshick_2014_CVPR,Redmon_2016_CVPR} lies in the bird's-eye view perspective of aerial images, which causes objects to appear in arbitrary orientations. If horizontal bounding boxes (commonly used in general object detection) are applied to aerial images, they may encompass excessive background and make it difficult to localize the target objects accurately. In recent years, several datasets \cite{HRSC,Xia_2018_CVPR, sun2022fair1m,DIOR} have been introduced that utilize oriented bounding boxes (OBB) for annotating aerial images, which can more tightly enclose the target objects. However, bounding boxes now include angle information, increasing the data dimensionality from 4 to 5 when this approach is applied, which makes detection more challenging. When objects appear in arbitrary orientations, the detector should produce consistent outputs for classification tasks, regardless of orientation, requiring rotation invariance. In contrast, the detector needs to correctly adjust the output angles, necessitating rotation equivariance for bounding box regression. Most aerial object detectors implicitly learn rotation equivariance by increasing model parameters \cite{9956816,DRN,9954050}, enhancing data augmentation \cite{OrientedRCNN,yang2021r3det,S2ANet,Yang_2019_SCRDet,RoITransformer, DODet,AOPG}, proposing new bounding box representation methods \cite{GlidingVertex,CenterMap,Grep,Oreppoints,ConvexHull,CSL} or introducing customized loss functions \cite{GWD,KLD,7560644,KFIoU}.

Cohen \textit{et al.} \cite{cohen-group-1} explicitly implement group-equivariant neural networks using group theory, which is further advanced in subsequent works such as \cite{cohen-steerable, E2CNN}. These developments lead to the creation of rotation-equivariant networks (RE-Nets) that are particularly well-suited for aerial object detection tasks. Some previous studies have applied RE-Nets to aerial object detection. ReDet \cite{ReDet} is the first to adopt a RE-Net as a backbone and neck. FRED \cite{FRED} combines the anchor-free point-set representation from RepPoints \cite{Yang_2019_ICCV} with RE-Nets to achieve a fully rotation-equivariant detector. Both ReDet and FRED achieved state-of-the-art performance at the time and have demonstrated the effectiveness of RE-Nets in aerial object detection. 

However, most object detectors use typical downsampling as a trade-off between computational efficiency and accuracy, performing 2x downsampling with stride-2 convolutional kernels during forward propagation. Additionally, the object detection pipeline often resizes image dimensions to multiples of two to facilitate mapping predicted bounding box coordinates back to the original image. Mohamed \textit{et al.} \cite{mohamed2020data} has shown that applying stride-2 convolutional kernels to feature maps with even spatial dimensions breaks strict rotation equivariance. Previous research on RE-Nets in aerial object detection has not explored how strict rotation equivariance impacts model performance. For instance, ReDet only achieves approximate rotation equivariance. FRED ensures strict rotation equivariance by padding even-dimensional feature maps with zeros to make them odd-dimensional before downsampling. However, this approach may introduce misalignment issues. 

This study proposes a new downsampling process that avoids single-sided padding. Instead, an additional tuning layer is introduced before the stride-2 downsampling layer to convert even-dimensional feature maps to odd dimensions to ensure strict rotation equivariance  (see \cref{fig:1}). Additionally, Edixhoven \textit{et al.} \cite{Using} have shown that approximately rotation-equivariant models may relax strict equivariance constraints during training in image classification tasks. In certain scenarios, both strict and approximate rotation equivariance have their advantages and disadvantages. However, the impact of strict versus approximate rotation equivariance on model performance in aerial detection tasks remains an open question. This study will control the strictness of rotation equivariance by determining whether to include tuning layers in the network, thus comparing the effects of strict and approximate rotation equivariance on model performance. And we track the rotation equivariance error during training of the approximately rotation-equivariant model on the aerial object detection dataset to assess the importance of rotation equivariance.

Among the state-of-the-art CNN-based object detectors \cite{CBAM, PPYOLOE} that have emerged in recent years, the channel attention mechanism proposed by SENet \cite{SENet} is a commonly used component. However, applying channel attention mechanisms in RE-Nets can break rotation equivariance. In this study, we implement a rotation-equivariant channel attention mechanism and develop a more advanced backbone and neck network based on RTMDet \cite{rtmdet}. Due to the inherent grouping properties of rotation-equivariant features, the proposed model inputs feature maps from different orientation dimensions into separate branches of the head network, which are then combined for output. Thanks to the higher degree of weight sharing in RE-Nets and the multi-branch head network, the model proposed in this paper has an exceptionally low number of parameters. To our knowledge, it is currently the model with the fewest parameters while maintaining state-of-the-art performance. The multi-branch head rotation-equivariant single-stage object detector (MessDet) we propose achieves advanced performance on DOTA-v1.0 \cite{DOTA}, DOTA-v1.5 \cite{DOTA} and DIOR-R \cite{DIOR}.

In summary, the contributions of this study are as follows:
\begin{itemize}[leftmargin=*, labelwidth=1.5em, labelsep=0.5em, itemindent=!, parsep=0pt]
    \item \hangindent=1.5em A novel downsampling process is proposed and integrated with RE-Nets, enabling control over whether the network exhibits strict or approximate rotation equivariance. 
    \item \hangindent=1.5em A rotation-equivariant channel attention mechanism is proposed to enable the implementation of rotation-equivariant backbone and neck networks with more advanced architectures. A multi-branch head network is introduced to make more efficient use of rotation-equivariant features.
    \item \hangindent=1.5em The variation of rotation equivariance error during training is quantitatively analyzed in the aerial detection task, highlighting the importance of rotation equivariance for aerial detection performance.
\end{itemize}

\section{Related Work}
\label{sec:Related Work}
This section briefly reviews the most pertinent studies and methodologies that have shaped the field with the aim of providing a foundation for the contributions of this work.
\subsection{Aerial Object Detection}

Due to the potential for objects in aerial images to appear at arbitrary angles, the detector’s output must include not only the bounding box position, width, height, and class as in general object detection but also the angle information of the bounding box. 
This distinction allows research on aerial image object detectors to be roughly divided into the following categories. First, some studies focus on improving network structures or adding specialized modules to enable the model to better learn features related to orientation and enhance rotation equivariance. R$^3$Det \cite{yang2021r3det} and DRN \cite{DRN} improve feature accuracy by adding feature refinement modules.
SCRDet \cite{Yang_2019_SCRDet} and SCRDet++ \cite{SCRDet++} introduce instance-level denoising modules to reduce interference from cluttered background regions. RoI Transformer \cite{RoITransformer} transforms the feature regions corresponding to horizontal anchor boxes into those corresponding to oriented boxes. Oriented R-CNN \cite{OrientedRCNN} generates OBB with lower computational cost. S$^2$ANet \cite{S2ANet} introduces AlignConv, which shifts convolutional kernels, and constructs a feature alignment network using AlignConv.
ReDet \cite{ReDet} is the first to introduce RE-Nets into aerial object detection tasks and improved RoI Align \cite{maskrcnn} by interpolating along the orientation dimension to align rotation-equivariant features, resulting in RiRoI Align. ARC \cite{ARC} proposes an adaptive rotation convolutional module that allows filters to rotate for objects at different orientations.
FRED \cite{FRED} re-applies RE-Nets and combines them with point-set representation and deformable convolutions. All of the above studies can be viewed as methods to make networks more robust to rotational changes. 
In addition, LSKNet \cite{LSKNet}, PKINet \cite{PKINet}, and Strip R-CNN \cite{striprcnn} tackle challenges such as background noise and extreme aspect ratios in aerial images by employing large kernel and strip convolutions. This approach significantly reduces the parameter count while greatly enhancing detection performance.

Other studies use alternative representations for oriented boxes. CenterMap OBB \cite{CenterMap} and Oriented RepPoints \cite{Oreppoints} use point-set representations for oriented boxes. Gliding Vertex \cite{GlidingVertex} represents oriented boxes based on horizontal boxes by regressing four length ratios. CFA \cite{ConvexHull} uses convex hulls to represent oriented boxes. GWD \cite{GWD} transforms the box representation into a two-dimensional Gaussian distribution, using Wasserstein Distance as the loss function. KLD \cite{KLD} is similar to GWD but uses Kullback-Leibler Divergence as the loss function, addressing the issue of scale invariance that GWD lacks. These representations eliminate the need to predict extra rotation angles, preventing training instability caused by drastic regression loss changes when large objects with small angular shifts alter their aspect ratios. Additionally, Cheng \textit{et al.} \cite{7560644} add an extra regularization term in the loss function to enforce the model's output features approximately equivariant to changes in image angles.

Similar to ReDet and FRED, this study aims to explicitly endow the model with rotation equivariance through RE-Nets, rather than enhancing the model's robustness to angular variations by designing additional feature alignment methods or bounding box representation techniques.

\subsection{Rotation-equivariant Networks}

Cohen \textit{et al.} \cite{cohen-group-1} first introduce group-equivariant neural networks and have achieved equivariance for the discrete rotation group. Harmonic Networks \cite{HarmonicNetworks} achieves 360-degree rotation equivariance by replacing conventional CNN filters with circular harmonics. Subsequent works \cite{cohen-steerable,Weiler_2018_CVPR} achieve better performance with convolutional kernels having fewer parameters. Spherical CNNs \cite{sphericalcnns} accelerates the group-equivariant convolution process using generalized fast Fourier transform, while E2CNN \cite{E2CNN} extends group-equivariant convolutional neural networks to the broader Euclidean group E(2) and experimentally demonstrate the performance advantages of E(2) convolutional kernels on multiple classification datasets. ReDet \cite{ReDet} is the first to apply RE-Nets to aerial object detection tasks, and FRED \cite{FRED} implements a fully rotation-equivariant network, both achieving state-of-the-art performance at the time, highlighting the effectiveness of RE-Nets in aerial object detection.

\section{Preliminaries}

This section introduces concepts related to rotation equivariance in this study.
\begin{figure}[t]
  \centering
   \includegraphics[width=0.80\linewidth]{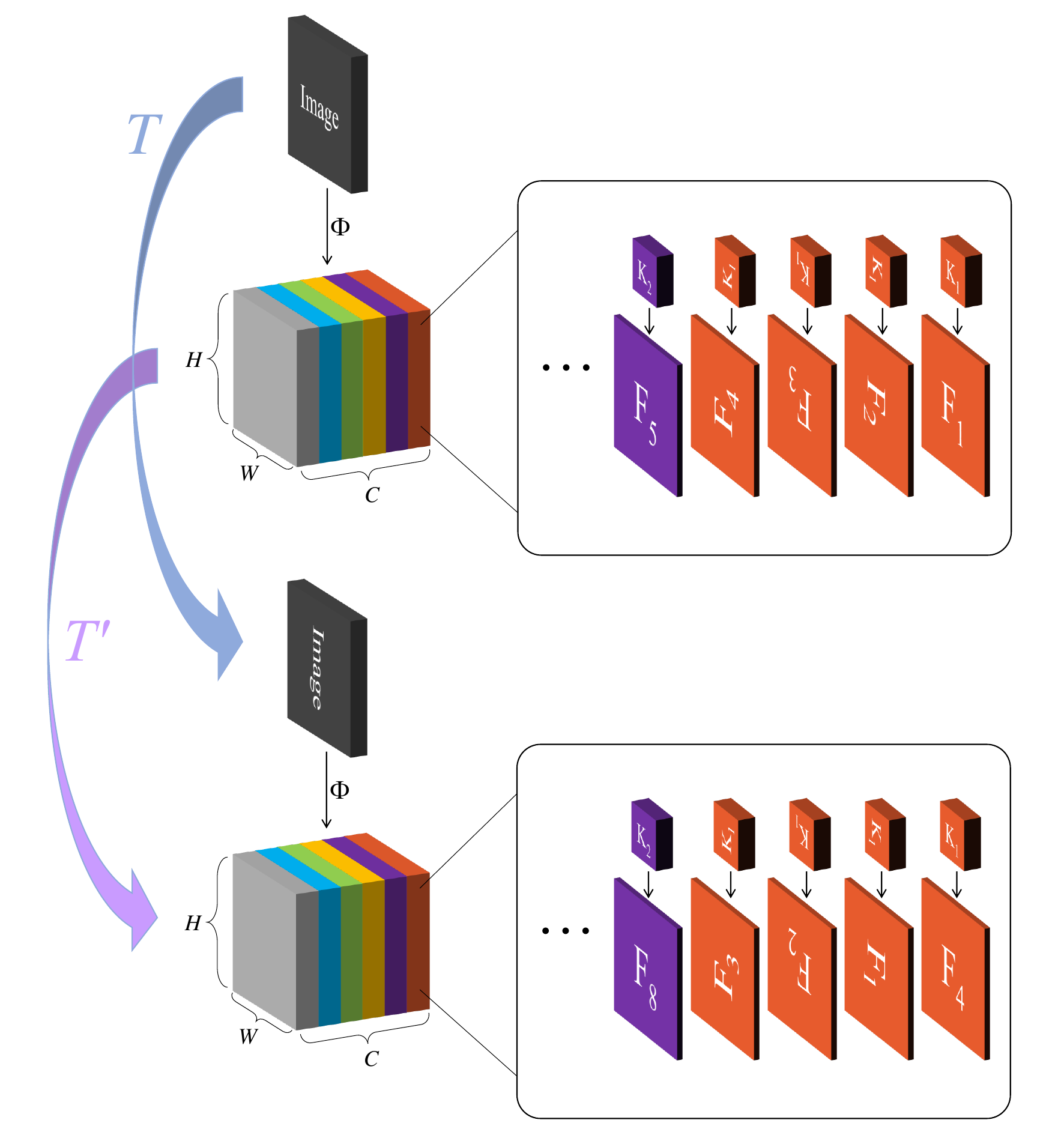}

   \caption{Rotation-equivariant feature. The upper half illustrates an input image processed by a RE-Net, which generates rotation-equivariant features. In the channel dimension, feature maps are shown in different colors, each representing a different kernel, such as K$_1$ (Kernel 1). The RE-Net applies K$_1$ at multiple orientations (four orientations are shown for illustration) and performs convolutions with the input feature maps to obtain different output feature maps, such as F$_1$ and F$_2$. The lower half illustrates the resulting feature map when the input image is first rotated 90$^\circ$ clockwise before being processed by the RE-Net. The convolutional kernels and their orientations remain unchanged, but F$_1$ shifts in the channel dimension and rotates 90$^\circ$ clockwise in the spatial dimension.}
   \label{fig:2}
\end{figure}

\subsection{Rotation Equivariance}

Let the signal be $x$, and after applying a certain transformation, it becomes $Tx$. If the output when $Tx$ is input into the network $\Phi$ should be the same as the output when $x$ is input directly into the network, after applying a certain transformation $T'$, that is,
\begin{equation}
  \Phi(Tx)=T'\Phi(x),
  \label{eq:1}
\end{equation}
the network $\Phi$ is equivariant to the transformation $T$ where $T$ and $T'$ do not need to be the same. In rotation-equivariant convolutional neural networks, $T$ is defined as a transformation on the group $G$. If $T'=1$, the network $\Phi$ is said to be invariant to the transformation $T$, so invariance is a special case of equivariance. Ordinary convolutional neural networks exhibit equivariance to translation but not to rotation \cite{cohen-group-1}. Rotation-equivariant convolutional neural networks (hereafter referred to as RE-Nets) achieve rotation equivariance by constructing cyclic groups, as shown in \cref{fig:2}.

In \cref{fig:2}, it can be observed that when the input image is rotated, the feature maps simultaneously shift within the cyclic group included in the channel dimension and undergo the corresponding rotation in the spatial dimension. So, $\Phi(T_g x)=T_g'\Phi(x)$, where $g$ represents an element of the group $G$. When the input is a rotation-equivariant feature, the convolutional kernel's rotation involves both spatial rotations and translations within the cyclic group in the channel dimension.

\subsection{Breaking Rotation Equivariance}
\label{section3.2}

In object detection tasks, it is common practice to repeatedly perform downsampling by a factor of 2, such as achieving 32$\times$ downsampling, to provide the network with a larger receptive field. Additionally, when mapping a certain region of the feature map back to the original image or converting coordinates on the feature map to original image coordinates, the image is often resized to a size that is a multiple of two to ensure feature alignment and precise boundary box localization. A typical downsampling method involves using a convolutional kernel with a stride of 2 on the feature map of even dimensions. However,  Mohamed \textit{et al.} \cite{mohamed2020data} points out that doing this can result in different sampling points for the convolutional kernel before and after rotation, which leads to breaking the rotation equivariance. A specific example can be found in Sec. 7 of the supplementary material.

In classification tasks, Edixhoven \textit{et al.} \cite{Using} experimentally demonstrated that for models that break rotation equivariance, the rotation equivariance error of shallow convolutional layers gradually decreases during the training process, resulting in approximate rotation equivariance. In some cases, the rotation equivariance error of deeper convolutional layers may even increase during training. Romero \textit{et al.} \cite{pmlr-v119-romero20a} use 2$\times$2 pooling for 2$\times$ downsampling to ensure rotation equivariance. However, this approach is not suitable for aerial object detection tasks, as 2$\times$2 pooling causes significant loss of information for small objects. In detection tasks, FRED \cite{FRED} uses padding to ensure that convolution with a stride of 2 always encounters odd-sized dimensions. However, single-side padding can also lead to misalignment of features, which may reduce the precision of the detector.

\section{The Design of MessDet}

\begin{figure}[t]
  \centering
   \includegraphics[width=0.99\linewidth]{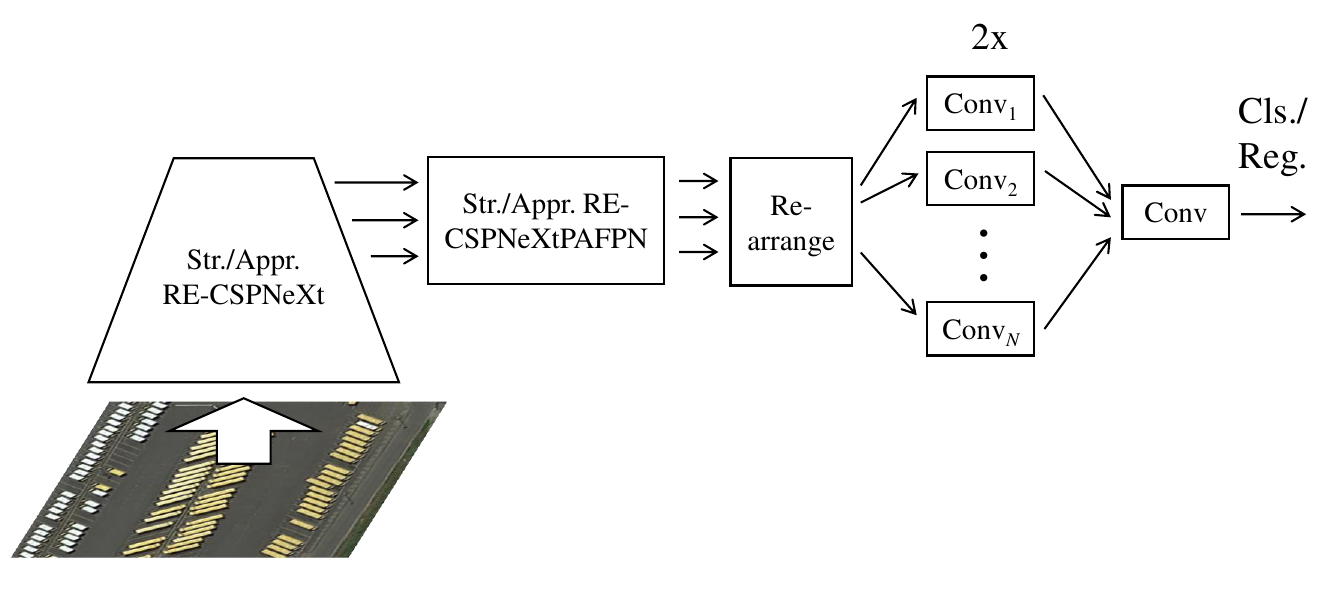}

   \caption{MessDet architecture. ``Str." and ``Appr." refer to strict rotation equivariance and approximate rotation equivariance, respectively. ``$N$" denotes the number of orientation dimensions.}
   \label{fig:3}
\end{figure}
This study uses RTMDet \cite{rtmdet} as the baseline and reimplements its backbone network CSPNeXt (an improved version of CSPNet \cite{CSPNet}) and neck network CSPNeXtPAFPN (an improved version of PANet \cite{PAN} and FPN \cite{FPN}), incorporating the E2CNN \cite{E2CNN}. Three improvements are made to handle rotation-equivariant features. These lead to the construction of a new aerial image object detector, MessDet. The network structure diagram is shown in \cref{fig:3}. The following subsections will detail the design of MessDet. 

\subsection{Downsampling While Keeping Rotation Equivariance}
\label{section4.1}

\begin{figure}
  \centering
  \begin{subfigure}{0.48\linewidth}
    \includegraphics[width=1\linewidth]{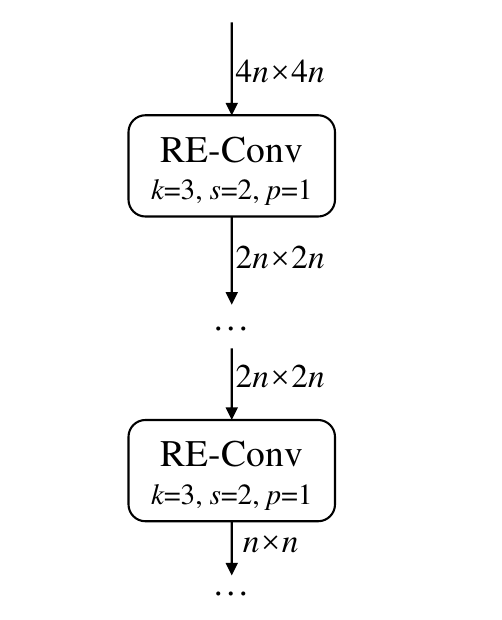}
    \caption{Common downsampling process.}
    \label{fig:4a}
  \end{subfigure}
  \hfill
  \begin{subfigure}{0.48\linewidth}
    \includegraphics[width=1\linewidth]{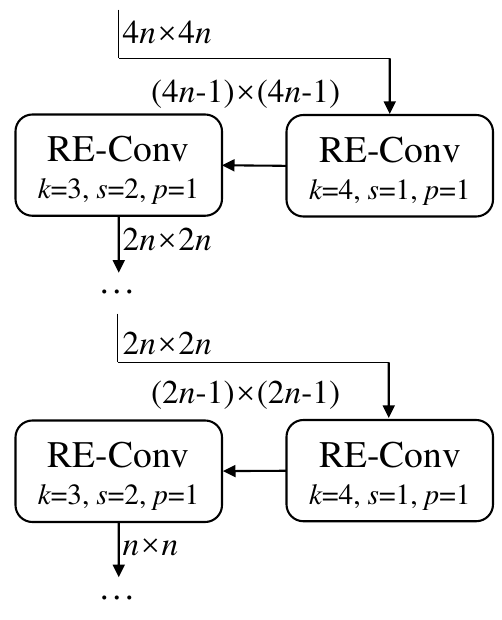}
    \caption{Strictly rotation-equivariant downsampling process.}
    \label{fig:4b}
  \end{subfigure}
  \caption{Comparison of different downsampling processes, where RE-Conv represents the rotation-equivariant convolutional module.}
  \label{fig:4}
\end{figure}

If a RE-Net directly uses a convolutional layer with a stride of 2 for downsampling on feature maps with even spatial dimensions, it will lead to different sampling points. As a result, the rotation equivariance is broken and only achieves approximate rotation equivariance, as shown in \cref{fig:4a}. This study proposes a novel downsampling process that maintains strict rotation equivariance, as shown in \cref{fig:4b}. Assuming that the width and height of the input feature map are the same, $S_{in}=H_{in}=W_{in}$, the relationship between the spatial dimensions of the output feature map from the convolutional layer and the input feature map is given by:
\begin{equation}
  S_{out}=\left\lfloor\frac{S_{in}+2p-d\cdot(k-1)-1}{s}\right\rfloor+1.
  \label{eq:7}
\end{equation}

In the Formula \ref{eq:7}, $S_{out}$ and $S_{in}$ represent the output size and input size, respectively, $p$ is the padding size, $d$ is the dilation (which is set to 1 in this study as dilation convolutions are not involved), $k$ is the kernel size, and $s$ is the stride of the convolutional kernel. To maintain strict rotation equivariance, feature maps with even spatial dimensions $2n$ are first passed through a convolutional layer with $k=4$, $p=1$ and $s=1$, referred to as the tuning layer. The output size of the tuning layer, calculated by Formula \ref{eq:7}, is $2n-1$. Then, it is passed through a downsampling convolutional layer with $k=3$, $p=1$, and $s=2$, and the output feature map size is given by: $S_{out}=\left\lfloor((2n-1)-1)/2\right\rfloor+1=n$.
It can be seen that the inclusion of the tuning layer does not change the output size of the downsampling layer, but it ensures that the input feature map size for the 2x downsampling convolutional layer is odd, thereby guaranteeing strict rotation equivariance.

Users can easily alter whether the model exhibits strict rotation equivariance by adding a tuning layer before the two-times downsampling convolutional layer.

\subsection{Rotation-Equivariant Channel Attention and Multi-branch Head}
\label{section4.2}

The channel attention mechanism introduced in SENet \cite{SENet} is used in many state-of-the-art object detectors \cite{CBAM,PPYOLOE,rtmdet}. However, directly applying channel attention to rotation-equivariant features would break rotation equivariance. Therefore, this study proposes a rotation-equivariant channel attention (RE-CA) mechanism. Channel attention re-weights convolutional kernels, suppressing unimportant channels and enhancing important ones. The rotation-equivariant convolutional layer generates $N$ convolutional kernels by rotating a single kernel in $N$ orientations. Only $C/N$ convolutional kernels are needed to generate a feature map with $C$ channels. Thus, the RE-CA only needs to produce $C/N$ weights, each of which is repeated $N$ times, to weight the channels of the feature map. This ensures the preservation of rotation equivariance. 

For a rotation-equivariant feature $\mathbf{X}\in {{\mathbb{R}}^{C\times H\times W}}$, it can also be written as $\mathbf{X}\in {{\mathbb{R}}^{\frac{C}{N}\times N\times H\times W}}$.The process of RE-CA is as follows: the spatial information of the $c$-th channel of \[\mathbf{X}=[{{\mathbf{x}}_{1}},{{\mathbf{x}}_{2}},...,{{\mathbf{x}}_{C}}]\] is compressed into a scalar ${{z}_{c}}$ using global average pooling, $c\in [1,C]$, resulting in the vector $\boldsymbol{z}=[{{z}_{1}},{{z}_{2}},...,{{z}_{C}}]$,
\begin{equation}
  {{z}_{c}}=\frac{1}{H\times W}\sum\limits_{i=1}^{H}{\sum\limits_{j=1}^{W}{{{\mathbf{x}}_{c}}(i,j)}}.
  \label{eq:9}
\end{equation}
Then, $\boldsymbol{z}$ is passed through a fully connected layer $\mathbf{W}\in {{\mathbb{R}}^{\frac{C}{N}\times C}}$ and an activation function $\sigma $,
\begin{equation}
  \boldsymbol{s}=\sigma (\mathbf{W}\cdot \boldsymbol{z}),
  \label{eq:10}
\end{equation}
in the weight vector $\boldsymbol{s}\in {{\mathbb{R}}^{\frac{C}{N}}}$. Each element of the weight vector is repeated $N$ times to obtain $\boldsymbol{{s}'}\in {{\mathbb{R}}^{C}}$, which is then multiplied with $\mathbf{X}$ channel-wise,
$
  {{\mathbf{\tilde{x}}}_{c}}={{{s}'}_{c}}\cdot {{\mathbf{x}}_{c}}
$,
to obtain the weighted rotation-equivariant feature $\mathbf{\tilde{X}}=[{{\mathbf{\tilde{x}}}_{1}},{{\mathbf{\tilde{x}}}_{2}},...,{{\mathbf{\tilde{x}}}_{C}}]$.

The rotation-equivariant features $\mathbf{X}\in {{\mathbb{R}}^{C\times H\times W}}$ consist of feature maps generated by $N$ convolutional kernels in different orientations, inherently exhibiting a grouping property. These features can be rearranged into $N$ groups based on the orientation of the convolutional kernels that generate them, $\mathbf{X}\in {{\mathbb{R}}^{N\times \frac{C}{N}\times H\times W}}$, with each group having $C/N$ channels. 
These groups are then fed into different branches of the head network, and the $N$ feature maps are concatenated for the final output. Thanks to the multi-branch design, the parameter count of the head network is significantly reduced. Further details of rotation-equivariant channel attention and feature rearrangement can be found in Sec. 8 of the supplementary material.

\section{Experiments}

\subsection{Datasets and Experiments Settings}

\begin{table*}[t]
  \centering
  \fontsize{8}{10}\selectfont
  \setlength{\tabcolsep}{3.5pt}
  \begin{tabular}{l|*{2}{c|}*{15}{c}} 
    \hline
    \textbf{Method} & \textbf{\#P$\downarrow$} & \textbf{mAP$\uparrow$} & \textbf{PL} & \textbf{BD} & \textbf{BR} & \textbf{GTF} & \textbf{SV} & \textbf{LV} & \textbf{SH} & \textbf{TC} & \textbf{BC} & \textbf{ST} & \textbf{SBF} & \textbf{RA} & \textbf{HA} & \textbf{SP} & \textbf{HC} \\
    \hline
    \multicolumn{18}{l}{\textbf{Two-stage}} \\
    \hline
    CenterMap\cite{CenterMap}&41.1M&71.59&89.02&80.56&49.41&61.98&77.99&74.19&83.74&89.44&78.01&83.52&47.64&65.93&63.68&67.07&61.59\\ 
    SCRDet\cite{Yang_2019_SCRDet}&41.9M&72.61&89.98&80.65&52.09&68.36&68.36&60.32&72.41&90.85&87.94&86.86&65.02&66.68&66.25&68.24&65.21\\ 
    RoI Trans.\cite{RoITransformer}&55.1M&74.05&89.01&77.48&51.64&72.07&74.43&77.55&87.76&90.81&79.71&85.27&58.36&64.11&76.50&71.99&54.06\\ 
    FR-Est\cite{FREst}&-&74.20&89.63&81.17&50.44&70.19&73.52&77.98&86.44&90.82&84.13&83.56&60.64&66.59&70.59&66.72&60.55\\ 
    G. Vertex\cite{GlidingVertex}&41.1M&75.02&89.64&85.00&52.26&77.34&73.01&73.14&86.82&90.74&79.02&86.81&59.55&\textbf{70.91}&72.94&70.86&57.32\\ 
    O-RCNN\cite{OrientedRCNN}&41.1M&75.87&89.46&82.12&54.78&70.86&78.93&83.00&88.20&\textbf{90.90}&87.50&84.68&63.97&67.69&74.94&68.84&52.28\\ 
    ReDet\cite{ReDet}&31.6M&76.25&88.79&82.64&53.97&74.00&78.13&84.06&88.04&90.89&87.78&85.75&61.76&60.39&75.96&68.07&63.59\\ 
    ARC\cite{ARC}&74.4M&77.35&89.40&82.48&55.33&73.88&79.37&84.05&88.06&\textbf{90.90}&86.44&84.83&63.63&70.32&74.29&71.91&65.43\\ 
    LSKNet\cite{LSKNet}&31.0M&77.49&89.66&\textbf{85.52}&\textbf{57.72}&75.70&74.95&78.69&88.24&90.88&86.79&86.38&66.92&63.77&77.77&74.47&64.82\\ 
    PKINet\cite{PKINet}&30.8M&78.39&\textbf{89.72}&84.20&55.81&\textbf{77.63}&80.25&84.45&88.12&90.88&87.57&86.07&66.86&70.23&77.47&73.62&62.94\\ 
    \hline
    \multicolumn{18}{l}{\textbf{Single-stage}} \\
    \hline
    R$^3$Det\cite{yang2021r3det}&41.9M&69.70&89.00&75.60&46.64&67.09&76.18&73.40&79.02&90.88&78.62&84.88&59.00&61.16&63.65&62.39&37.94\\ 
    DRN\cite{DRN}&-&70.70&88.91&80.22&43.52&63.35&73.48&70.69&84.94&90.14&83.85&84.11&50.12&58.41&67.62&68.60&52.50\\ 
    S$^2$ANet\cite{S2ANet}&38.5M&74.12&89.11&82.84&48.37&71.11&78.11&78.39&87.25&90.83&84.90&85.64&60.36&62.60&65.26&69.13&57.94\\ 
    SASM\cite{SASM}&36.6M&74.92&86.42&78.97&52.47&69.84&77.30&75.99&86.72&90.89&82.63&85.66&60.13&68.25&73.98&72.22&62.37\\ 
    FRED\cite{FRED}&-&75.56&89.37&82.12&50.84&73.89&77.58&77.38&87.51&90.82&86.30&84.25&62.54&65.10&72.65&69.55&63.41\\ 
    G-Rep\cite{Grep}&36.6M&75.56&87.76&81.29&52.64&70.53&80.34&80.56&87.47&90.74&82.91&85.01&61.48&68.51&67.53&73.02&63.54\\ 
    O-RepPoints\cite{Oreppoints}&36.6M&75.97&87.02&83.17&54.13&71.16&80.18&78.40&87.28&\textbf{90.90}&85.97&86.25&59.90&70.49&73.53&72.27&58.97\\ 
    R3Det-GWD\cite{GWD}&41.9M&76.34&88.82&82.94&55.63&72.75&78.52&83.10&87.46&90.21&86.36&85.44&64.70&61.41&73.46&76.94&57.38\\ 
    R3Det-KLD\cite{KLD}&41.9M&77.36&88.90&84.17&55.80&69.35&78.72&84.08&87.00&89.75&84.32&85.73&64.74&61.80&76.62&78.49&\textbf{70.89}\\ 
    RTMDet\cite{rtmdet}&52.3M&78.85&89.43&84.21&55.20&75.06&80.81&84.53&88.97&\textbf{90.90}&87.38&87.25&63.09&67.87&78.09&80.78&69.13\\ 
    \hline
    \rowcolor{gray!20}
    Appr. MessDet&\textbf{15.3M}&78.45&89.28&84.69&56.20&66.18&81.37&85.33&88.86&90.78&88.28&87.62&64.47&65.59&78.18&82.10&67.82\\ 
    \rowcolor{gray!20}
    Str. MessDet&18.1M&\textbf{79.12}&88.08&85.15&55.81&72.09&\textbf{81.54}&\textbf{85.85}&\textbf{88.98}&90.84&\textbf{88.36}&\textbf{88.70}&\textbf{68.78}&66.85&\textbf{78.84}&\textbf{82.66}&64.26\\ 
    \hline
  \end{tabular}
  \caption{\textbf{Comparison with state-of-the-art methods on the DOTA-v1.0 dataset} \cite{Xia_2018_CVPR} with single-scale training and testing. The MessDet backbones are pretrained on ImageNet-1K \cite{ImageNet} for 300 epochs, similar to other methods \cite{LSKNet, PKINet,yang2021r3det}.}
  \label{tab:DOTA-v1.0}
\end{table*}
We evaluate models on DOTA-v1.0 \cite{Xia_2018_CVPR,DOTA}, DOTA-v1.5 and DIOR-R \cite{DIOR}. DOTA-v1.0 contains 2,806 aerial images with OBB annotations. The resolutions of these images range from 800×800 to 4000×4000. It includes 188,282 instances across 15 categories: Plane (PL), Baseball diamond (BD), Bridge (BR), Ground track field (GTF), Small vehicle (SV), Large vehicle (LV), Ship (SH), Tennis court (TC), Basketball court (BC), Storage tank (ST), Soccer-ball field (SBF), Roundabout (RA), Harbor (HA), Swimming pool (SP), and Helicopter (HC). The image data in DOTA-v1.5 is the same as that in DOTA-v1.0, but it includes instances smaller than 10 pixels and a new category, Container Crane (CC). DOTA-v1.5 contains 402,089 instances. DIOR-R \cite{DIOR} includes 23,463 aerial images and 192,518 annotations across 20 categories.

\begin{table}[t]
  \centering
  \begin{tabular}{c|c|c|c}
    \textbf{Method} & \textbf{Strictly} & \textbf{\#P$\downarrow$} & \textbf{mAP$\uparrow$} \\
    \hline
    \multirow{2}{*}{\makecell{MessDet\\\small(Without head)}} & \ding{51} & 19.0M & 78.51 \\
                               & \ding{55} & 16.2M & 78.15 \\
    \hline
    \multirow{2}{*}{\makecell{RTMDet\\\small(Without EMA)}}  & \ding{51} & 74.9M & 77.10 \\
                               & \ding{55} & 52.3M & 77.34 \\
  \end{tabular}
  \caption{Performance comparison of RTMDet and MessDet with two downsampling methods.}
  \label{tab:1}
\end{table}

For a fair comparison, we follow the approach of major methods \cite{S2ANet, OrientedRCNN}, training models on both the training and validation sets and evaluating them on the test set. We adopt single-scale training and testing, using 1024×1024 patches for DOTA and 800×800 resized images for DIOR-R.
Similar to mainstream methods \cite{LSKNet,yang2021r3det}, we pretrain the strictly and approximately rotation-equivariant backbone networks used in the main results on ImageNet-1K \cite{ImageNet} for 300 epochs, while the backbones in the ablation studies are pretrained for 100 epochs. Our model is implemented using the MMYOLO \cite{mmyolo2022} and MMRotate \cite{zhou2022mmrotate} frameworks and trained for 36 epochs on DOTA-v1.0, DOTA-v1.5 and DIOR-R using the AdamW \cite{AdamW} optimizer. The experiments are conducted on 4 GPUs with a batch size of 8. Following the exploration of ReDet \cite{ReDet}, this study set $N$, the number of orientation dimensions for the rotation-equivariant features in MessDet, to 8. More detailed experimental settings can be found in Sec. 9 of the supplementary material.

\begin{table*}[t]
  \centering
  \fontsize{8}{10}\selectfont
  \setlength{\tabcolsep}{2.8pt}
  \begin{tabular}{l|*{2}{c|}*{16}{c}} 
    \hline
    \textbf{Method} & \textbf{\#P$\downarrow$} & \textbf{mAP$\uparrow$} & \textbf{PL} & \textbf{BD} & \textbf{BR} & \textbf{GTF} & \textbf{SV} & \textbf{LV} & \textbf{SH} & \textbf{TC} & \textbf{BC} & \textbf{ST} & \textbf{SBF} & \textbf{RA} & \textbf{HA} & \textbf{SP} & \textbf{HC} & \textbf{CC}\\
    \hline
    \multicolumn{19}{l}{\textbf{Two-stage}} \\
    \hline
    FR-O\cite{fasterrcnn}&41.1M&62.00&71.89&74.47&44.45&59.87&51.28&68.98&79.37&90.78&77.38&67.50&47.75&69.72&61.22&65.28&60.47&1.54\\ 
    Mask R-CNN\cite{maskrcnn}&44.4M&62.67&76.84&73.51&49.90&57.80&51.31&71.34&79.75&90.46&74.21&66.07&46.21&70.61&63.07&64.46&57.81&9.42\\ 
    HTC\cite{HTC}&77.5M&63.40&77.80&73.67&51.40&63.99&51.54&73.31&80.31&90.48&75.12&67.34&48.51&70.63&64.84&64.48&55.87&5.15\\ 
    RoI Trans.\cite{RoITransformer}&55.1M&65.50&71.70&82.70&53.00&71.50&51.30&74.60&80.60&90.40&78.00&68.30&53.10&73.40&73.90&65.60&56.90&3.00\\ 
    ReDet\cite{ReDet}&31.6M&66.86&79.20&82.81&51.92&71.41&52.38&75.73&80.92&90.83&75.81&68.64&49.29&72.03&73.36&70.55&63.33&11.53\\ 
    LSKNet\cite{LSKNet}&31.0M&70.26&72.05&84.94&55.41&\textbf{74.93}&52.42&77.45&81.17&90.85&79.44&69.00&62.10&\textbf{73.72}&77.49&75.29&55.81&\textbf{42.19}\\ 
    PKINet\cite{PKINet}&30.8M&71.47&80.31&85.00&\textbf{55.61}&74.38&52.41&76.85&88.38&\textbf{90.87}&79.04&68.78&\textbf{67.47}&72.45&76.24&74.53&64.07&37.13\\ 
    
    \hline
    \multicolumn{19}{l}{\textbf{Single-stage}} \\
    \hline
    
    RetinaNet-O\cite{retinanet}&36.4M&59.16&71.43&77.64&42.12&64.65&44.53&56.79&73.31&90.84&76.02&59.96&46.95&69.24&59.65&64.52&48.06&0.83\\ 
    O-RepPoints\cite{Oreppoints}&36.6M&66.90&75.52&82.60&51.24&70.21&57.81&73.82&86.25&90.86&78.30&76.47&53.61&72.78&66.68&69.48&53.66&11.09\\ 
    FRED\cite{FRED}&-&68.30&79.60&81.44&52.60&72.57&58.07&74.82&86.12&90.81&82.13&74.84&53.37&72.93&69.51&69.91&54.82&19.27\\ 
    RTMDet\cite{rtmdet}&52.3M&71.80&80.00&80.75&51.82&71.65&67.12&80.69&89.48&90.76&81.43&81.07&60.87&70.41&74.87&73.29&\textbf{70.90}&23.65\\ 
    \hline
    \rowcolor{gray!20}
    Appr. MessDet&\textbf{15.3M}&72.38&80.61&84.99&54.56&65.36&73.60&82.69&89.73&90.83&82.46&82.03&59.35&72.96&\textbf{77.66}&75.14&65.09&21.12\\ 
    \rowcolor{gray!20}
    Str. MessDet&18.1M&\textbf{73.14}&\textbf{80.83}&\textbf{86.17}&55.58&63.28&\textbf{74.24}&\textbf{82.73}&\textbf{89.79}&90.85&\textbf{83.47}&\textbf{82.45}&63.97&71.65&77.44&\textbf{76.06}&66.17&25.54\\ 
    \hline
  \end{tabular}
  \caption{\textbf{Comparison with state-of-the-art methods on the DOTA-v1.5 dataset} \cite{DOTA} with single-scale training and testing. The MessDet backbones are pretrained on ImageNet-1K \cite{ImageNet} for 300 epochs, similar to other methods \cite{PKINet,ReDet}.}
  \label{tab:DOTA-v1.5}
\end{table*}

\subsection{Ablation Studies}

This section presents ablation experiments to validate the effectiveness of the newly proposed downsampling process, rotation-equivariant channel attention (RE-CA) mechanism, and multi-branch head network. Unless otherwise stated, the reported results are based on the DOTA-v1.0 test set.

\textbf{Strictly Rotation-Equivariant Downsampling.}
We have implemented rotation-equivariant backbone and neck networks with both approximately and strictly rotation-equivariant downsampling. To compare the impact of the new downsampling process on conventional convolutional networks, we re-implement RTMDet \cite{rtmdet} with both approximately and strictly equivariant downsampling processes. For a fair comparison, MessDet uses the same head network as RTMDet, and RTMDet does not employ the  Exponential Moving Average (EMA) strategy. Note that we only pretrain RTMDet's two backbones for 100 epochs instead of the 600 epochs as in the original paper. The new downsampling process has little impact on conventional CNN but improves MessDet implemented with RE-Nets, as shown in \cref{tab:1}.

\textbf{Rotation-Equivariant Channel Attention.}
To demonstrate the effectiveness of the proposed RE-CA mechanism, we implement strictly and approximately rotation-equivariant backbones without the RE-CA, and the head network without the multi-branch structure. The experimental results are shown in \cref{tab:2}. The results indicate that the model with RE-CA performs significantly better than the model without it. Furthermore, due to the maintenance of rotation equivariance, the output dimension of the fully connected layer is reduced to $1/N$ of the original. Hence, the parameter count required by the new channel attention mechanism is lower than that of conventional channel attention.
\begin{table}[t]
  \centering
  \begin{tabular}{c|c|c|c}
    \textbf{Strictly} & \textbf{RE-CA} & \textbf{\#P$\downarrow$} & \textbf{mAP$\uparrow$} \\
    \hline
    \multirow{2}{*}{\ding{51}} & \ding{51} & 19.0M & \textbf{78.51} \\
                               &  \ding{55} & 18.8M & 76.91 \\ 
    \hline
    \multirow{2}{*}{\ding{55}}  & \ding{51} & 16.2M & \textbf{78.15} \\
                               & \ding{55} & 16.0M & 77.47 \\
  \end{tabular}
  \caption{Performance comparison of MessDet with and without rotation-equivariant channel attention.}
  \label{tab:2}
\end{table}

\textbf{Multi-branch Head.}
We conduct a series of experiments to demonstrate that the multi-branch head network not only reduces the number of parameters but also improves model performance. To achieve the best possible performance, this study explores the impact of the number of convolutional modules in the multi-branch network on performance, using the Appr. MessDet as an example, as shown in \cref{tab:3}. The number of convolutional modules includes the number of convolutional modules in each individual branch plus one convolutional module for the output aggregation. This shows that when the number of convolutional modules in the multi-branch head is set to 3, a balance between performance and parameter count is achieved. 
\begin{table}[t]
  \centering
  \begin{tabular}{c|c|c|c}
    \textbf{} & \textbf{Num.} & \textbf{\#P\small(head only)\normalsize$\downarrow$} & \textbf{mAP$\uparrow$} \\
    \hline
    \multirow{1}{*}{RTMDet Head} & 2 & 2.4M & 78.15 \\
    \hline
    \multirow{5}{*}{Multi-branch Head}  & 9 & 2.4M & 78.19 \\
                                & 7 & 2.1M & 77.97 \\
                                & 5 & 1.8M & 77.62 \\
                                & 3 & 1.5M & \textbf{78.45} \\
                                & 2 & 1.4M & 77.46 \\
  \end{tabular}
  \caption{The impact of different head network designs on the performance of Appr. MessDet.}
  \label{tab:3}
\end{table}

\subsection{Main Results}

\newcolumntype{C}{>{\centering\arraybackslash}X}
\begin{table}[t]
  \centering
  \fontsize{9}{11}\selectfont 
  \begin{tabularx}{\linewidth}{l | C C}
    \hline
    \textbf{Method} & \textbf{\#P$\downarrow$} & \textbf{mAP$\uparrow$}\\
    \hline
    RetinaNet OBB\cite{retinanet}&36.4M&57.55\\ 
    Faster R-CNN OBB\cite{fasterrcnn}&41.1M&59.54\\ 
    RoI Transformer\cite{RoITransformer}&55.1M&63.87\\ 
    LSKNet\cite{LSKNet}&31.0M&65.90\\ 
    Oriented Reppoints\cite{Oreppoints}&36.6M&66.71\\ 
    DCFL\cite{DCFL}&-&66.80\\ 
    RTMDet\cite{rtmdet}&52.3M&66.86\\ 
    PKINet\cite{PKINet}&30.8M&67.03\\ 
    \hline
    \rowcolor{gray!20}
    Appr. MessDet&\textbf{15.3M}&67.42\\ 
    \rowcolor{gray!20}
    Str. MessDet&18.1M&\textbf{68.19}\\ 
    \hline
  \end{tabularx}
  \caption{\textbf{Comparison with state-of-the-art methods on the DIOR-R dataset} \cite{DIOR}. The MessDet backbones are pretrained on ImageNet-1K \cite{ImageNet} for 300 epochs, similar to other methods \cite{Oreppoints,PKINet,DCFL}. }
  \label{tab:DIOR}
\end{table}

\textbf{On DOTA-v1.0} \cite{DOTA}, the performance comparison between MessDet and other state-of-the-art methods is reported in \cref{tab:DOTA-v1.0}. The approximately rotation-equivariant MessDet (Appr. MessDet) achieves 78.45\% mAP, while the strictly rotation-equivariant MessDet (Str. MessDet) achieves 79.12\% mAP. However, compared to other state-of-the-art models, MessDet has nearly half the parameter count of the lowest among them. 

\noindent \textbf{On DOTA-v1.5} \cite{DOTA}, \cref{tab:DOTA-v1.5} presents the results, showing that Appr. MessDet attains 72.38\% mAP, while Str. MessDet achieves 73.14\% mAP.

\noindent \textbf{On DIOR-R} \cite{DIOR}, as detailed in \cref{tab:DIOR}, Appr. MessDet records an mAP of 67.03\%, whereas Str. MessDet improves upon this with 68.19\% mAP.

\subsection{Analysis}

This section will analyze the behavior of rotation-equivariant networks in aerial object detection tasks, aiming to demonstrate the importance of rotation equivariance in such tasks.

\textbf{Equivariance Error Reduction in Detectors.}
\begin{figure}[t]
  \centering
   \includegraphics[width=0.99\linewidth]{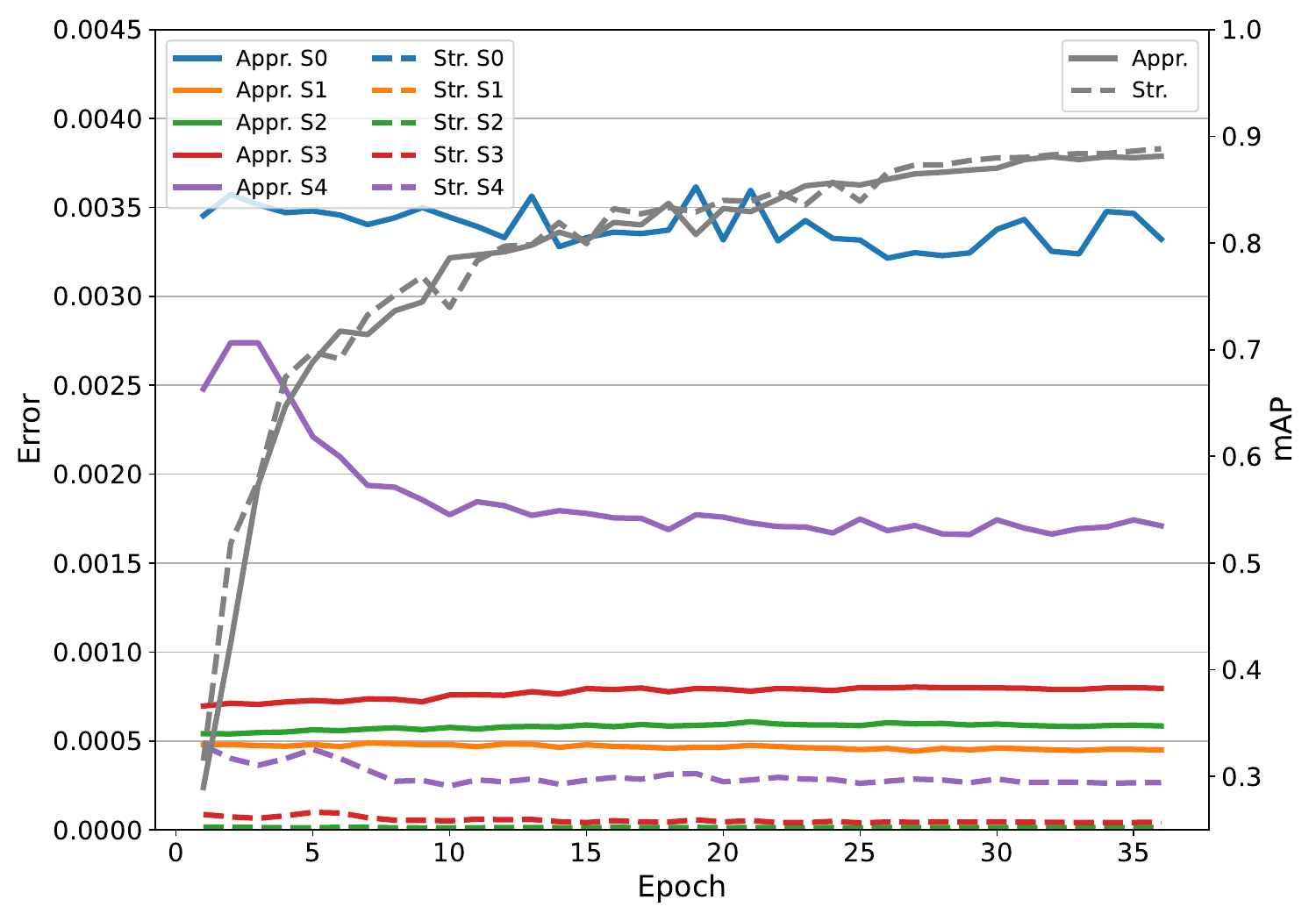}

   \caption{The variation of equivariance error across the five stages of the MessDet backbone network, where “Appr.” and “Str.” denote approximately and strictly rotation-equivariant MessDet respectively, and S0 denotes the stem, S1 denotes stage 1, and so on. mAP represents the model's performance on the DOTA-v1.0 validation set.}
   \label{fig:error}
\end{figure}
Edixhoven \textit{et al.} \cite{Using} point out that in classification tasks, approximate rotation-equivariant models relax the rotation equivariance constraint, gradually increasing the rotation equivariance error during training to improve performance. Following this paper, we define the rotation equivariance error as follows:
\begin{equation}
  \varepsilon =\frac{1}{ijk}\sqrt{\sum\limits_{i,j,k}{{{({{[f(T(x))]}_{ijk}}-{{[{T}'(f(x))]}_{ijk}})}^{2}}}}.
  \label{eq:error}
\end{equation}
In Eq.(\ref{eq:error}), $i$ and $j$ are indices in the spatial dimensions, and $k$ is the index in the channel dimension. To eliminate possible extraneous interference, we separately test the Appr. and Str. MessDet models without the multi-branch head network. The variation in rotation equivariance error across the five stages of the backbone network is shown in \cref{fig:error}, based on the DOTA-v1.0 dataset.

As shown in \cref{fig:error}, unlike in classification tasks, the rotation equivariance error in the approximately rotation-equivariant model decreases gradually during training on the DOTA-v1.0 aerial object detection dataset, proving that aerial image object detection tasks indeed benefit more from rotation equivariance. Furthermore, the rotation equivariance error in the strictly RE-Net is naturally lower than in the approximately RE-Net, leading to improved detection performance.

\begin{figure}[t]
  \centering
   \includegraphics[width=0.99\linewidth]{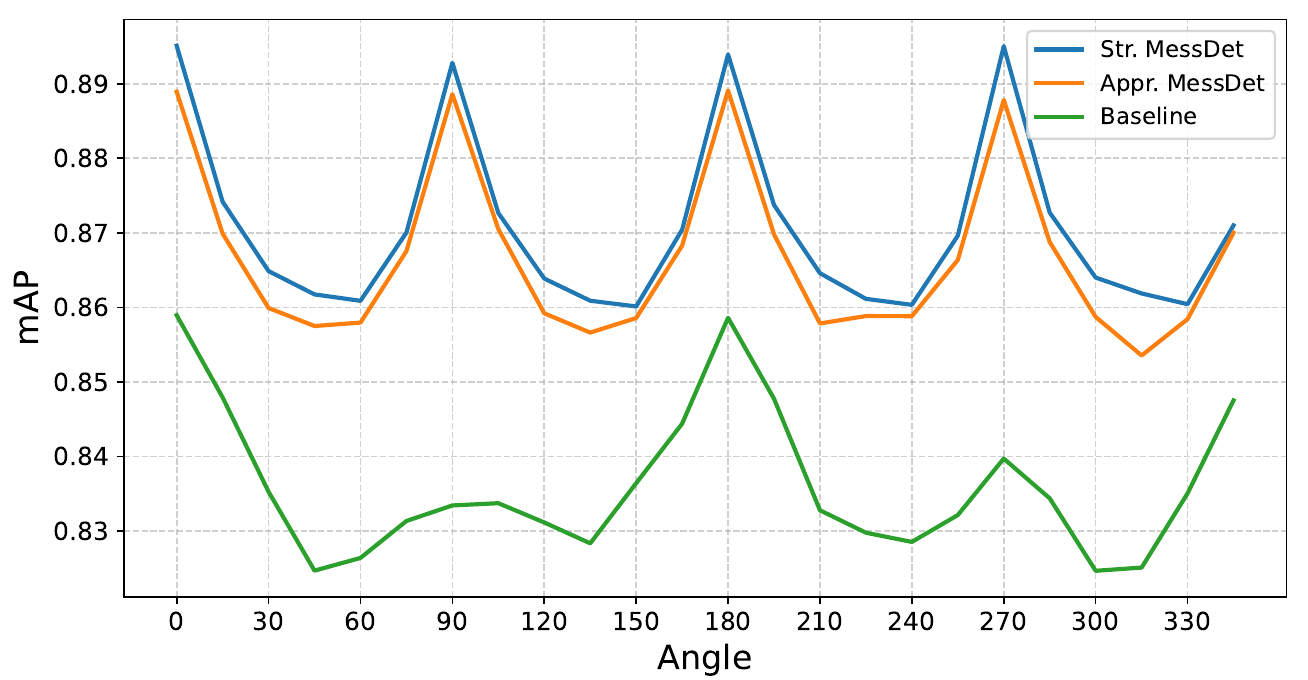}

   \caption{The impact of various rotation angles on models performance.}
   \label{fig:robust}
\end{figure}

\textbf{Robustness to Rotational Variations.}
To demonstrate the robustness of RE-Nets to rotational variations, we test both Str. and Appr. MessDet, along with the baseline RTMDet \cite{rtmdet}, by applying rotations at various angles to the DOTA-v1.0 validation set. The effect of rotational changes on model accuracy is compared. Similarly, to eliminate any unnecessary interference, MessDet uses the same head network as RTMDet. The test results are shown in \cref{fig:robust}. As can be seen, even when RTMDet is rotated by multiples of 90$^\circ$ with no information loss, its performance decreases due to a lack of robustness to rotational variations, whereas MessDet exhibits greater robustness to such variations.

\section{Conclusion}

This paper proposes a novel downsampling process, implements a strictly rotation-equivariant network, and introduces a rotation-equivariant channel attention mechanism, enabling rotation-equivariant networks to adopt more advanced architectures. By combining a multi-branch head network, we present the MessDet. Experimental results show that we achieve performance comparable to state-of-the-art methods with an exceptionally low parameter count. Additionally, we quantitatively evaluate rotation equivariance error, demonstrating the necessity of rotation equivariance in aerial image object detection tasks.

\vskip 0.2 cm
\noindent\textbf{Acknowledgement.} This work was supported in part by the National Natural Science Foundation of China (Nos. 62302364, 62306324, 62376279, U24A20333), the Science and Technology Innovation Program of Hunan Province (No. 2024RC3128), the Research Foundation of National University of Defense Technology (No. ZK24-30), and the Key Laboratory of Target Cognition and Application Technology (No. 2023-CXPT-LC-005).

{
    \small
    \bibliographystyle{ieeenat_fullname}
    \bibliography{main}
}

\clearpage
\setcounter{page}{1}
\maketitlesupplementary

\section{An Example Where Downsampling Breaks Rotation Equivariance}
\label{sec:7}

\begin{figure}[t]
	\centering
	\includegraphics[width=0.99\linewidth]{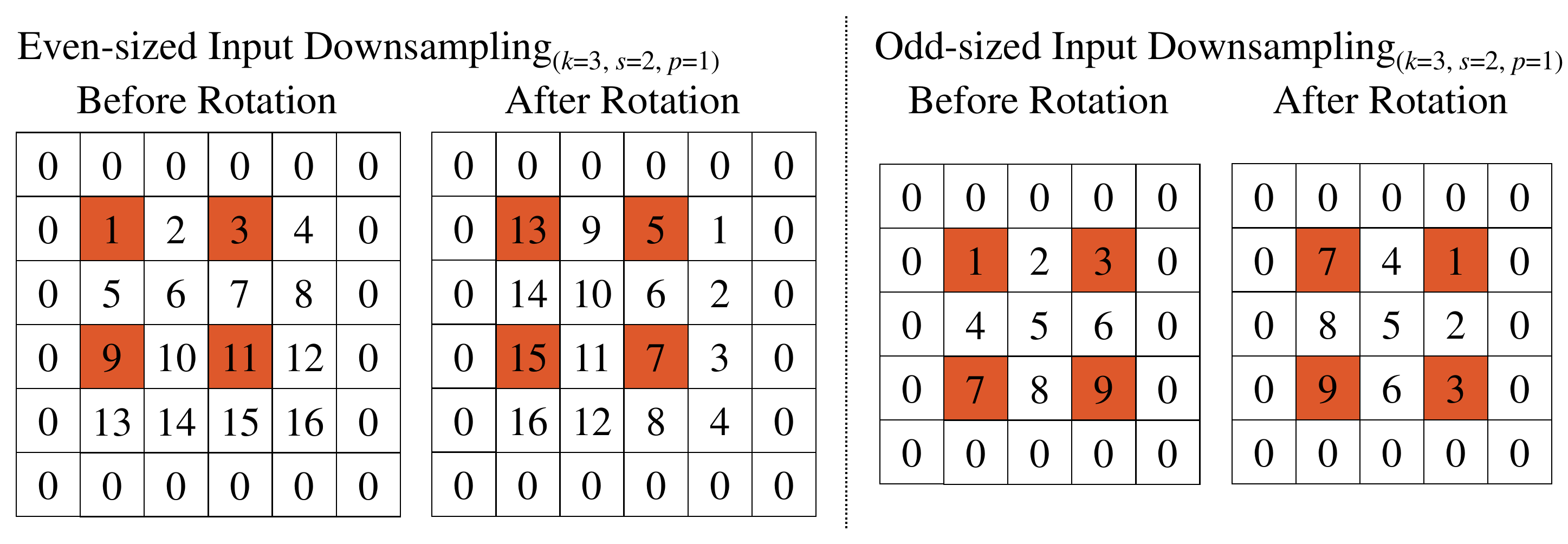}
	\caption{Breaking and maintaining strict rotation equivariance.}
	\label{fig:visualized grid}
\end{figure}

In Section \ref{section3.2}, we describe how conventional downsampling layers can break strict rotation equivariance. This occurs because the center sampling points of the convolution kernels do not match before and after rotation on even-sized feature maps, as illustrated in the left part of Fig. \ref{fig:visualized grid}. The grid illustrates the padded feature map and the orange-highlighted pixel represents the convolution kernel's center sampling point.

For example, consider a clockwise 90-degree rotation. Define a grayscale image as a matrix, assuming the image has a size of $2n\times2n$. Let $x$ represent the column index and $y$ represent the row index, $1\le x,y\le 2n,x,y\in\mathbb{Z}$. When performing downsampling on the image using a convolutional kernel with a stride of 2, the coordinates of the sampling points are as follows:
\begin{equation}
  (x,y)=(2i+1,2j+1),i,j\in[0,n-1].
  \label{eq:3}
\end{equation}
If $f$ represents the grayscale value of a pixel in the image at a certain coordinate, after rotating the image 90 degrees clockwise, the relationship between the coordinates $(x',y')$ of the rotated pixel and the original pixel coordinates is given by:
\begin{equation}
  f(x',y')=f(2n-y+1,x).
  \label{eq:4}
\end{equation}
Similarly, after rotating the image 90 degrees clockwise, if a stride-2 convolution kernel is used for downsampling, the coordinates of the sampling points in the rotated image are given by:
\begin{equation}
  (x',y')=(2i+1,2j+1),i,j\in[0,n-1].
  \label{eq:5}
\end{equation}
By substituting the corresponding relationships from \ref{eq:4} into \ref{eq:5}, the coordinates of the sampling pixel points in the rotated image can be expressed in terms of their original coordinates as:
\begin{equation}
  (x,y)=(2j+1,2n-2i),i,j\in[0,n-1].
  \label{eq:6}
\end{equation}
From the differences between Eq.(\ref{eq:3}) and Eq.(\ref{eq:6}), it can be observed that the row indices of the sampling points before rotation are all odd, while the row indices of the sampling points after rotation are all even. The difference in sampling points leads to varying convolution results for the same convolution kernel on the same feature map, which breaks strict rotation equivariance.

To address this issue, we ensure strict rotation equivariance by using odd-sized inputs for all 2x downsampling layers, which preserves the alignment of convolution kernel center points before and after rotation, as illustrated in the right part of Fig. \ref{fig:visualized grid}.
The theoretical proof, as shown in Section 3.2 of ~\cite{Using}, establishes that, to maintain strict rotation equivariance, the input size $i$, kernel size $k$, and stride $s$ of a downsampling layer—regardless of whether it is a convolutional layer or a pooling layer—must satisfy the condition $(i - k) \bmod s = 0$. Notably, the downsampling method proposed in this study adheres to this condition.

\section{The Further Details of MessDet}

\begin{table*}[t]
	\centering
	\fontsize{8}{10}\selectfont
	\setlength{\tabcolsep}{3.5pt}
	\begin{tabular}{l|*{2}{c|}*{15}{c}} 
		\hline
		\textbf{Method} & \textbf{\#P$\downarrow$} & \textbf{mAP$\uparrow$} & \textbf{PL} & \textbf{BD} & \textbf{BR} & \textbf{GTF} & \textbf{SV} & \textbf{LV} & \textbf{SH} & \textbf{TC} & \textbf{BC} & \textbf{ST} & \textbf{SBF} & \textbf{RA} & \textbf{HA} & \textbf{SP} & \textbf{HC} \\
		\hline
		\multicolumn{18}{l}{\textbf{Two-stage}} \\
		\hline
		SCRDet\cite{Yang_2019_SCRDet}&41.9M&72.61&89.98&80.65&52.09&68.36&68.36&60.32&72.41&90.85&87.94&86.86&65.02&66.68&66.25&68.24&65.21\\ 
		CSL\cite{CSL}&37.4M&76.17&\textbf{90.25}&85.53&54.64&75.31&70.44&73.51&77.62&90.84&86.15&86.69&69.60&68.04&73.83&71.10&68.93\\ 
		ReDet\cite{ReDet}&31.6M&80.10&88.81&82.48&60.83&80.82&78.34&86.06&88.31&90.87&\textbf{88.77}&87.03&68.65&66.90&79.26&79.71&74.67\\ 
		DODet\cite{DODet}&-&80.62&89.96&85.52&58.01&81.22&78.71&85.46&88.59&90.89&87.12&87.80&70.50&\textbf{71.54}&82.06&77.43&74.47\\ 
		AOPG\cite{AOPG}&-&80.66&89.88&85.57&60.90&81.51&78.70&85.29&88.85&90.89&87.60&87.65&71.66&68.69&82.31&77.32&73.10\\ 
		LSKNet\cite{LSKNet}&31.0M&\textbf{81.64}&89.57&86.34&\textbf{63.13}&\textbf{83.67}&\textbf{82.20}&\textbf{86.10}&88.66&90.89&88.41&87.42&71.72&69.58&78.88&81.77&76.52\\ 
		
		\hline
		\multicolumn{18}{l}{\textbf{Single-stage}} \\
		\hline
		
		R$^3$Det\cite{yang2021r3det}&41.9M&76.47&89.80&83.77&48.11&66.77&78.76&83.27&87.84&90.82&85.38&85.51&65.57&62.68&67.53&78.56&72.62\\ 
		CFA\cite{ConvexHull}&-&76.67&89.08&83.20&54.37&66.87&81.23&80.96&87.17&90.21&84.32&86.09&52.34&69.94&75.52&80.76&67.96\\ 
		SASM\cite{SASM}&-&79.17&89.54&85.94&57.73&78.41&79.78&84.19&\textbf{89.25}&90.87&58.80&87.27&63.82&67.81&78.67&79.35&69.37\\ 
		S$^2$Net\cite{S2ANet}&-&79.42&88.89&83.60&57.74&81.95&79.94&83.19&89.11&90.78&84.87&87.81&70.30&68.25&78.30&77.01&69.58\\ 
		R$^3$Det-GWD\cite{GWD}&41.9M&80.23&89.66&84.99&59.26&82.19&78.97&84.83&87.70&90.21&86.54&86.85&\textbf{73.47}&67.77&76.92&79.22&74.92\\ 
		RTMDet\cite{rtmdet}&52.3M&80.54&88.36&84.96&57.33&80.46&80.58&84.88&88.08&\textbf{90.90}&86.32&87.57&69.29&70.61&78.63&80.97&\textbf{79.24}\\ 
		R$^3$Det-KLD\cite{KLD}&41.9M&80.63&89.92&85.13&59.19&81.33&78.82&84.38&87.50&89.80&87.33&87.00&72.57&71.35&77.12&79.34&78.68\\ 
		
		\hline
		\rowcolor{gray!20}
		Appr. MessDet&\textbf{15.3M}&80.36
		&88.45&85.50&59.10&81.51&79.97&84.49&88.32&90.89&87.39&87.20&69.55&69.63&78.16&\textbf{81.96}&73.23\\ 
		\rowcolor{gray!20}
		Str. MessDet&18.1M&81.07
		&88.40&\textbf{86.54}&60.84&82.71&81.41&85.64&88.99&90.89&88.58&\textbf{88.05}&71.42&68.41&\textbf{83.78}&80.94&69.50\\ 
		\hline
	\end{tabular}
	\caption{\textbf{Comparison with state-of-the-art methods on the DOTA-v1.0 dataset} \cite{DOTA} with multi-scale training and testing. The mAP in parentheses refers to the COCO-style mAP.}
	\label{tab:DOTA-v1.0ms}
\end{table*}

\begin{figure}[t]
  \centering
   \includegraphics[width=0.99\linewidth]{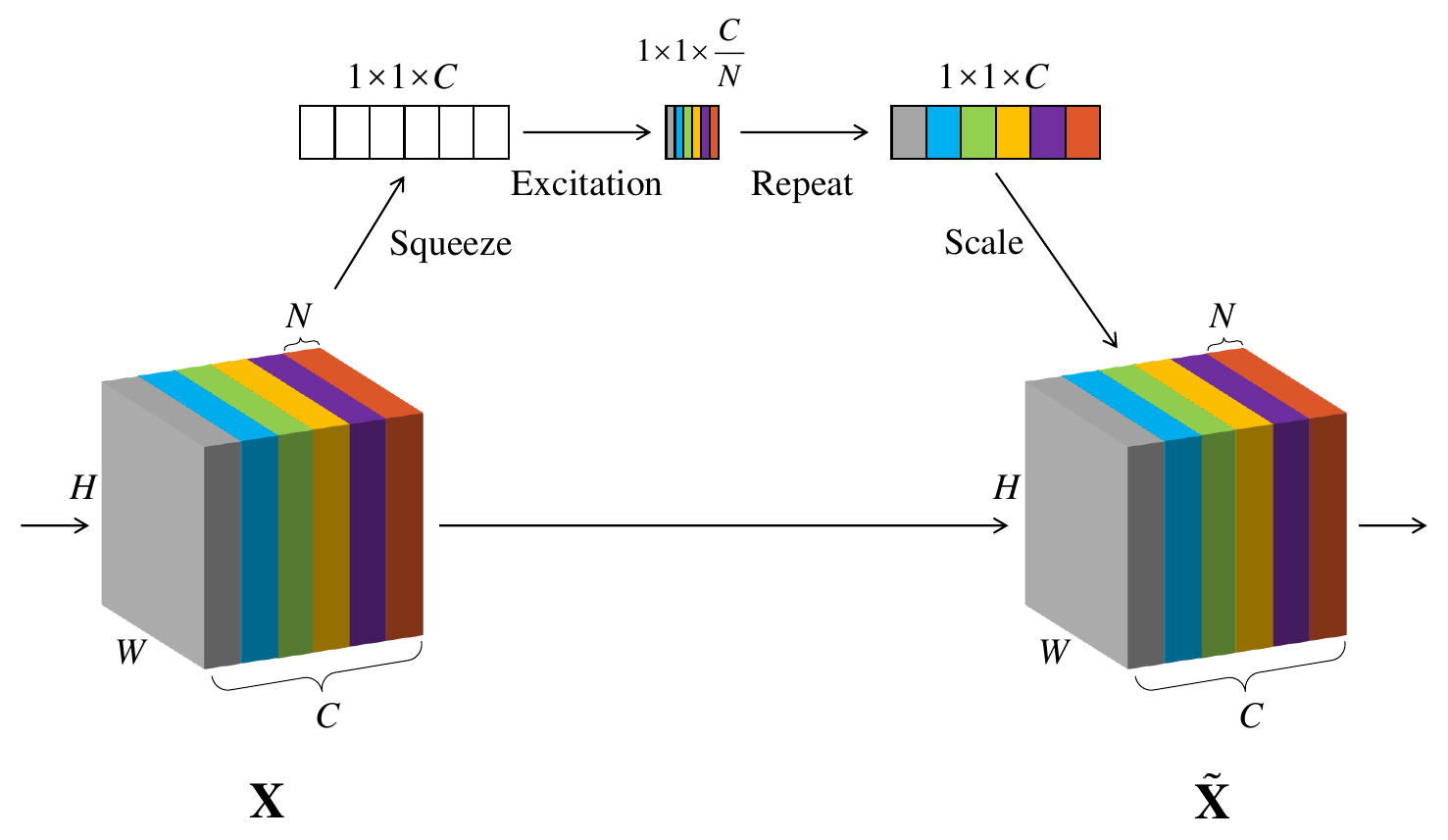}

   \caption{Rotation-Equivariant Channel Attention Mechanism.}
   \label{fig:5}
\end{figure}

\label{sec:8}
This paper introduces the rotation-equivariant channel attention (RE-CA), enabling rotation-equivariant networks to be implemented with more advanced network structures. The mathematical formulation of RE-CA is provided in Section \ref{section4.2}, and its schematic diagram is shown in Fig. \ref{fig:5}. In the figure, "Squeeze" refers to the global average pooling process, and "Excitation" refers to the fully connected layer and activation function. After obtaining the $C/N$-dimensional weight vector, each component of the vector is repeated $N$ times to obtain the $N$-dimensional weight vector that preserves rotation equivariance.

\begin{figure}[t]
  \centering
   \includegraphics[width=0.85\linewidth]{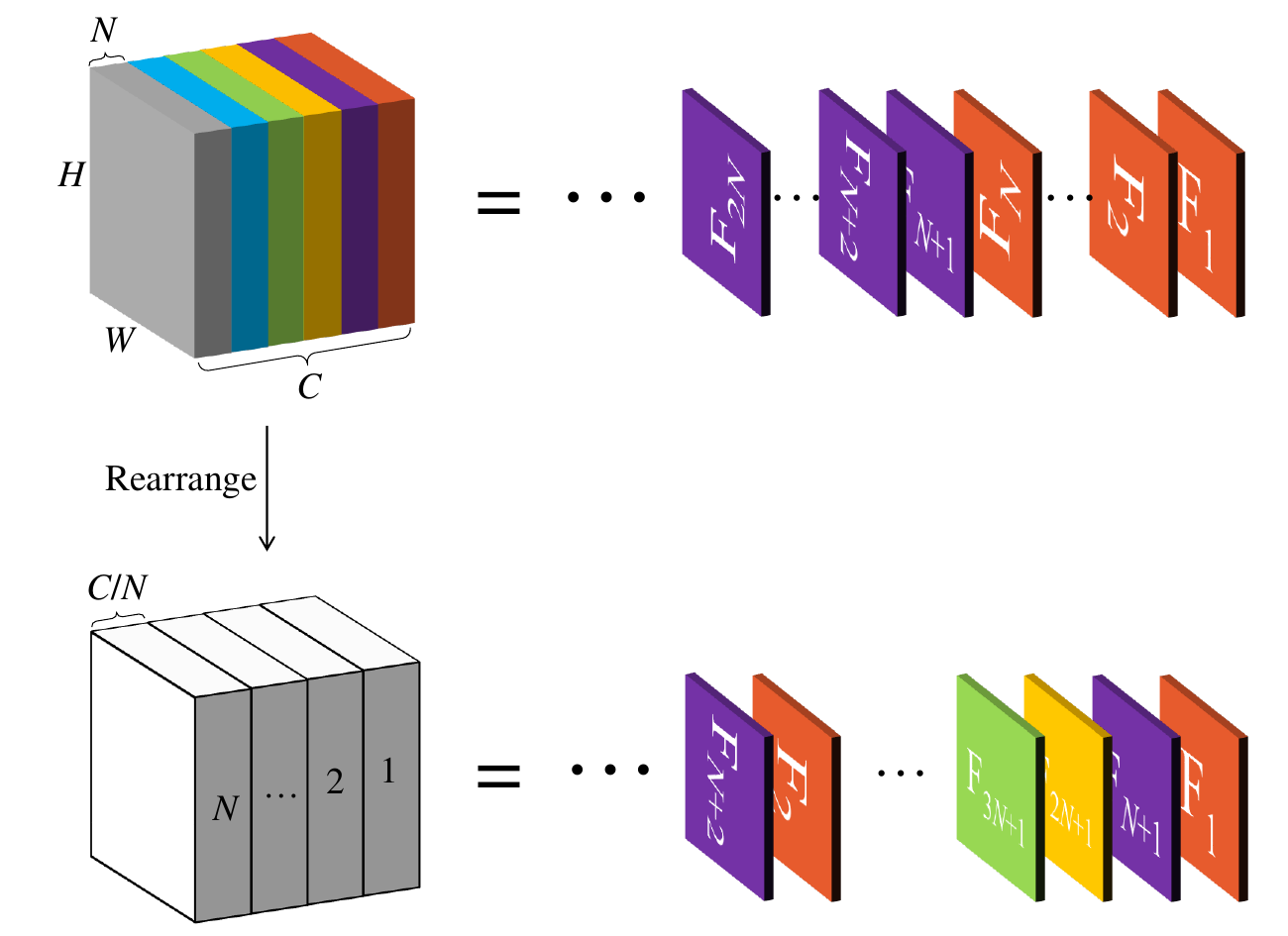}

   \caption{Illustration of Feature Rearrangement.}
   \label{fig:6}
\end{figure}

In the head network of MessDet, features with inherent group properties from different orientations are fed to different branches. To achieve this, the features are rearranged along the channel dimension. The rearrangement process is shown in Fig. \ref{fig:6}. For a rotation-equivariant feature, $\mathbf{X}\in {{\mathbb{R}}^{C\times H\times W}}$, where the feature map index of a certain channel is $c$, channels with the same remainder when dividing $c$ by $N$ are grouped together, thus grouping the channels generated by convolution kernels in different orientations.

\section{Experiments Details and The Further Experiments}

\begin{table}[t]
	\centering
	\fontsize{8.5}{10}\selectfont
	\begin{tabular}{c|c|c|c}
		\textbf{MessDet\scalebox{0.85}{(without head)}} & \textbf{Strictly} & \textbf{mAP} & \textbf{COCO-mAP} \\
		\hline
		\multirow{2}{*}{\makecell{on DOTA-v1.0}}  & \ding{51} & \textbf{78.51} & \textbf{51.96} \\
		& \ding{55} & 78.15 & 51.58 \\
		\hline
		\multirow{2}{*}{\makecell{on DOTA-v1.5}} & \ding{51} & \textbf{72.42} & \textbf{46.02} \\
		& \ding{55} & 71.26 & 43.54 \\ 
		
	\end{tabular}
	\caption{Performance comparison of MessDet with two downsampling methods on DOTA-v1.0 and DOTA-v1.5.}
	\label{tab:further1}
\end{table}

\begin{table}[t]
	\centering
	\fontsize{8.0}{10}\selectfont
	\begin{tabular}{c|c|c|c}
		\textbf{Method} & \textbf{FLOPs(G)} & \textbf{FPS (img/s)} & \textbf{Training Time(H)}  \\
		\hline
		Str. MessDet & 570 & 25.4 & 10.5 \\
		Appr. MessDet & 378 & 38.2 & 7.7 \\
	\end{tabular}
	\caption{Information on Inference Speed, Training Time, and FLOPs.}
	\label{tab:FLOPs}
\end{table}

\label{sec:9}
Our model is implemented using the MMYOLO ~\cite{mmyolo2022} and MMRotate ~\cite{zhou2022mmrotate} frameworks and trained for 36 epochs on DOTA-v1.0, DOTA-v1.5 and DIOR-R.
During training, we followed most mainstream methods ~\cite{PKINet, ReDet, OrientedRCNN,LSKNet} by employing random rotation and random flipping to prevent over-fitting.
The AdamW \cite{AdamW} optimizer is used with a base learning rate of 0.00025, weight decay of 0.05, and momentum of 0.9. The learning rate is gradually reduced to 1/20 of the base learning rate over the last half epochs using a cosine learning schedule. The experiments are conducted on 4 RTX 3090 GPUs with a batch size of 8. Following the exploration of ReDet \cite{ReDet} and FRED \cite{FRED}, this study set $N$, the number of orientation dimensions for the rotation-equivariant features in MessDet, to 8. 

Here we adopt both single-scale and multi-scale training strategies. For single-scale training and testing on DOTA-v1.0 and DOTA-v1.5, we crop the original images into 1024×1024 patches with a stride of 824, yielding a pixel overlap of 200 between adjacent patches. For multi-scale training and testing, we first resize the original images to three scales (0.5, 1.0, and 1.5), and then crop them into 1024×1024 patches with a stride of 524, resulting in an overlap of 500 pixels.

\textbf{Further Ablation Study on Rotation-Equivariant Downsampling.}
To further verify the performance gains brought by strictly equivariant downsampling, we conducted ablation experiments on the DOTA-v1.5 dataset using MessDet without the multi-branch head, and reported the COCO-style mAP, as shown in Table \ref{tab:further1}.
It can be observed that, since the model performance on DOTA-v1.0 has reached a bottleneck in recent years, Str. MessDet shows only marginal improvement over its approximate counterpart. However, on the more challenging DOTA-v1.5 dataset, Str. MessDet achieves a significantly larger performance gain compared to Appr. MessDet.

\textbf{Main Results with Multi-Scale Training on DOTA-v1.0.}
We also conducted multi-scale training experiments on the DOTA-v1.0 dataset for reference. Str. MessDet outperforms Appr. MessDet by 0.7 mAP, as shown in Tab. \ref{tab:DOTA-v1.0ms}

\textbf{Information on Inference Speed, Training Time, and FLOPs.} 
We provide the relevant metrics in Tab. \ref{tab:FLOPs} for reference, based on DOTA-v1.0 with 4 RTX 3090 GPUs for single-scale training and a single RTX 3090 GPU for inference. The FLOPs are calculated based on an input image size of 1024×1024.

\end{document}